\pdfoutput=1

\documentclass[11pt]{article}

\usepackage[]{acl}
\usepackage{times}
\usepackage{latexsym}

\usepackage[T1]{fontenc}

\usepackage[utf8]{inputenc}

\usepackage{microtype}

\usepackage{inconsolata}
\usepackage{microtype}
\usepackage{enumitem}
\usepackage{amssymb}
\usepackage{mathrsfs}
\usepackage{amsmath}
\usepackage{hhline}
\usepackage{amsfonts}
\usepackage{multirow}
\usepackage{array}
\usepackage{booktabs}
\usepackage{colortbl}
\usepackage{color}
\usepackage{subcaption}
\usepackage{threeparttable}
\usepackage{float}
\usepackage{caption}
\usepackage{graphicx}
\usepackage{arydshln}
\usepackage{makecell}
\usepackage{inconsolata}
\usepackage{rotating}
\usepackage{xcolor}
\usepackage{hyperref}
\usepackage{placeins}
\usepackage{longtable}
\usepackage[framemethod=TikZ]{mdframed}
\usepackage{hhline}
\definecolor{mygray}{gray}{0.9}
\definecolor{mypink}{rgb}{0.99,0.91,0.95}
\definecolor{mycyan}{cmyk}{0.3,0,0,0}
\definecolor{Celadon}{RGB}{172, 225, 175}
\definecolor{Peach}{RGB}{255, 229, 180}
\title{RST-LoRA: A Discourse-Aware Low-Rank Adaptation for Long Document Abstractive Summarization}

\author{Dongqi Liu \quad Vera Demberg \\
        Department of Computer Science\\
        Department of Language Science and Technology\\
        Saarland Informatics Campus, Saarland University, Germany\\
        \texttt{\{dongqi,vera\}@lst.uni-saarland.de}}

\begin{document}
\maketitle
\begin{abstract}
For long document summarization, discourse structure is important to discern the key content of the text and the differences in importance level between sentences. Unfortunately, the integration of rhetorical structure theory (RST) into parameter-efficient fine-tuning strategies for long document summarization remains unexplored. Therefore, this paper introduces \textbf{RST-LoRA} and proposes four RST-aware variants to explicitly incorporate RST into the LoRA model. Our empirical evaluation demonstrates that incorporating the type and uncertainty of rhetorical relations can complementarily enhance the performance of LoRA in summarization tasks. Furthermore, the best-performing variant we introduced outperforms the vanilla LoRA and full-parameter fine-tuning models, as confirmed by multiple automatic and human evaluations, and even surpasses previous state-of-the-art methods\footnote{The project information is available at \url{https://dongqi.me/projects/RST-LoRA}.}.
\end{abstract}

\section{Introduction}
The advent of pre-trained large language models (LLMs), such as LLaMA-2 \cite{touvron2023llama}, Vicuna \cite{zheng2023judging}, and GPT-related models from OpenAI \cite{OpenAI2023GPT4TR}, has greatly accelerated research progress of Natural Language Processing (NLP). With the continual growth in the scale of LLMs, the requirements for both software and hardware in order to fully fine-tune LLMs to adapt to downstream tasks, especially in processing long sequence data, will become increasingly demanding \cite{gu-etal-2022-ppt, pu2024scinews}.

Parameter-Efficient Fine-Tuning (PEFT) strategies are noteworthy in mitigating the aforementioned problem by reducing the number of parameters that need to be adjusted \cite{chen-etal-2022-revisiting, akbartajari-etal-2022-empirical, mao-etal-2022-unipelt, gheini-etal-2023-know, badola-etal-2023-parameter, zhang-etal-2023-towards-adaptive, lawton-etal-2023-neural}. Some studies have highlighted that by updating only 0.01--1\% of the (additional) parameters and freezing all other parameters of LLMs, PEFT methods can match or even exceed the performance of vanilla full-parameter fine-tuning \cite{li-liang-2021-prefix, hu2022lora, asai-etal-2022-attempt, yang-etal-2022-parameter, gu-etal-2023-gradient, liao-etal-2023-parameter, zhang-etal-2023-towards-adaptive, li-etal-2023-prefix, lei2023conditional, zhang2023adaptive, chen2023parameterefficient, lawton-etal-2023-neural}. Among these methods, LoRA algorithm \cite[Low-Rank Adaptation,][]{hu2022lora} has achieved state-of-the-art (SOTA) performance due to its ability to circumvent the latency associated with adapter tuning \cite{pmlr-v97-houlsby19a} as well as the input length constraints of prefix/prompt tuning \cite{li-liang-2021-prefix, lester-etal-2021-power} during model training and inference \cite{he2022towards, ghazvininejad-etal-2022-discourse, dettmers2023qlora, zhang2023adaptive, Whitehouse2023ParameterEfficientMS, ding-etal-2023-sparse}.

Recent investigations \cite{ustun-cooper-stickland-2022-parameter, ponti-etal-2023-combining, zhao-etal-2023-infusing, zeng-etal-2023-one, zhang-etal-2023-towards-adaptive, wan-etal-2023-pip, liu-etal-2023-recap} have revealed that PEFT strategies face challenges in distinguishing latent text relations and determining the importance level of different sentences during fine-tuning. This issue arises because such distinctions are not a primary focus in PEFT's learning process and are not explicitly represented in the input data. However, this is essential for the task of long document summarization since generating a good summary often requires natural language generation (NLG) models to have the ability to discern salient information within the text and comprehend the intricate interrelations among different text components. 

Our approach proposed here takes inspiration from \citet{ishigaki-etal-2019-discourse, xiao-etal-2020-really, dong-etal-2021-discourse, cao-wang-2022-hibrids, guo-etal-2022-modeling, pu-etal-2023-incorporating}, who have advised that explicitly integrating document structure and/or discourse knowledge can enhance the performance of neural summarization models when fully fine-tuning the NLG models. This motivates us to investigate the following research questions: \textit{Can the Rhetorical Structure Theory} \citep[\textit{RST},][]{mann1987rhetorical} \textit{improve the performance of LoRA strategy in summarizing lengthy documents?} Specifically, we want to explore and verify whether infusing RST knowledge into LoRA can improve the performance of long document summarization. To answer this question, this paper will propose, introduce, and integrate four RST structure variants to guide the training of LoRA. These variants include $(\romannumeral1)$ binary $(\romannumeral2)$ probability RST distribution, both with (w) and without(w/o) relation labels.

\textbf{In summary, our contributions are as follows:}

\begin{itemize}[leftmargin=8pt,itemsep=1pt,topsep=1pt,parsep=1pt]
    \item We introduce a method for injecting discourse knowledge into the training of the LoRA model. Our approach is compatible with both Seq2Seq and GPT transformer-based architectures, allowing for its easy adoption across different LLMs.
    \item Our empirical findings demonstrate that discourse uncertainty and relation labels are complementary, and both can contribute to the improvement of final performance. Notably, our model also outperforms current SOTA full-parameter fine-tuning (FFT) models in specific evaluation metrics.
    \item We offer quantitative and qualitative analyses showing that our model surpasses baseline models in factual consistency checking. Moreover, the results of human evaluation and GPT-4 examination reveal that our model produces summaries that are closer in quality to those generated by humans.
\end{itemize}

\section{RST Prerequisite Knowledge}

Rhetorical Structure Theory (RST) is a discourse framework that is helpful for determining which sentences in a document should or should not be included in a summary \cite{marcu1997discourse, marcu1999discourse, marcu2000theory, kikuchi-etal-2014-single, goyal-eisenstein-2016-joint, pu-etal-2023-incorporating}. To be specific, RST delineates a set of coherence relations between text segments, known as Elementary Discourse Units (EDUs), at the document level (e.g., one segment might provide clarification for another, or conversely, two segments could present contrasting viewpoints). Moreover, RST categorizes EDUs based on their discourse importance, labeling central EDUs as `nuclei', and less central EDUs as `satellites' \cite{marcu1999discourse, bosselut-etal-2018-discourse, isonuma-etal-2019-unsupervised, xu-etal-2020-discourse}. 

\begin{figure}[t]
  \centering
  \includegraphics[width=0.4\textwidth]{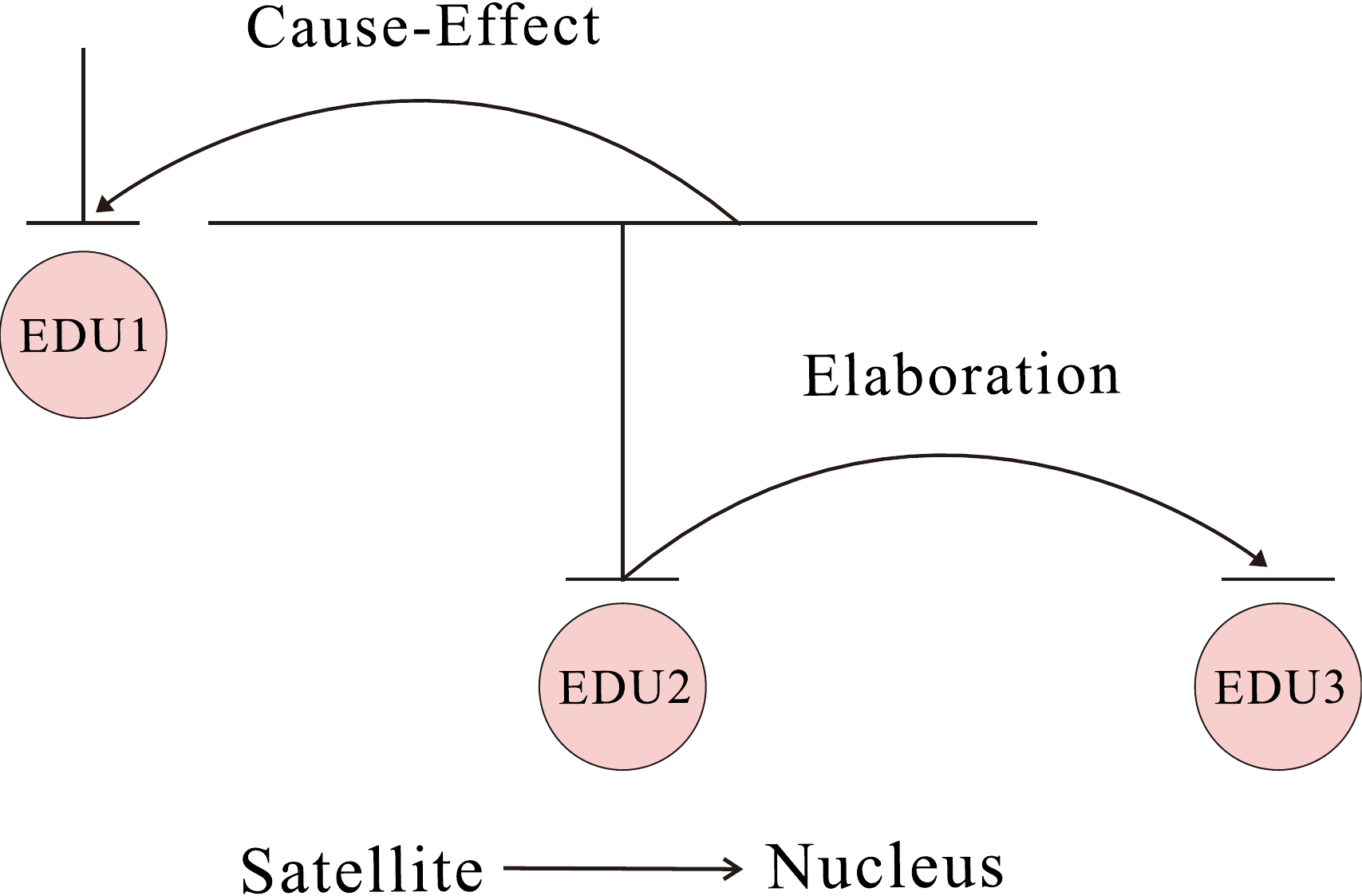}
  \caption{An example of RST tree: [\textit{Utilizing discourse structure to enhance text summarization is beneficial.}]$^{\mathrm{EDU1}}$ [\textit{This technique can be used to identify key ideas and capture often overlooked nuances.}]$^{\mathrm{EDU2}}$ [\textit{Accurate capture of these complex structures facilitates the generation of good summaries.}]$^{\mathrm{EDU3}}$}
  \label{fig:RST_sample}
\end{figure}

For example, consider an RST tree as depicted in Figure \ref{fig:RST_sample}. In this instance, EDU1 serves as the most pivotal component within the entire example, thus constituting the nucleus for both EDU2 and EDU3. EDU2 is tasked with elucidating and providing supplementary information to EDU3, positioning it as a satellite unit in relation to EDU3. Given the relatively diminished importance of EDU2, merging EDU1 and EDU3 while pruning EDU2, the semantic essence of the example would remain intact. In a more extreme scenario, retaining only EDU1 as the summary sentence and omitting both EDU2 and EDU3, the primary information conveyed by the entire example would still preserved. As has also been argued by, e.g., \citet{marcu1997discourse, louis-etal-2010-discourse, cohan-etal-2018-discourse, liu-etal-2019-single, li-etal-2020-composing, xu-etal-2020-discourse, dong-etal-2021-discourse, chen-yang-2021-structure} that `satellite' EDUs play a subordinate role in summarization, with most summary sentences deriving from `nuclei'.

\section{Related Work}

\subsection{Document Summarization with RST}

RST is a linguistic discourse framework that provides a way to organize text into a hierarchical tree structure, which helps to better understand the overall organization and inter-part relations of text. Early research by \citet{marcu1997discourse} and \citet{louis-etal-2010-discourse} uncovered that human-written summaries often align with the nucleus EDUs in RST trees. This correlation underscores the validity of RST as a theoretical motivation for summarization tasks. Building on this insight, subsequent studies have demonstrated the value of explicitly incorporating RST trees into neural summarization models. For example, \citet{kikuchi-etal-2014-single} boosted the summarization performance of the RNN-based model by constructing RST trees, where satellite EDUs were pruned to retain only the nucleus EDUs, thus focusing on the document's key content. Pre-trained language models have a noted tendency to capture some superficial aspects of discourse relations without explicit training \cite{miaschi-etal-2020-linguistic, qian-etal-2021-structural, schuster-linzen-2022-sentence}, but the latent discourse information is often not captured correctly. To alleviate this challenge, \citet{xu-etal-2020-discourse} and \citet{dong-etal-2021-discourse} enhanced summarization models by incorporating discourse structure within transformer-based and graph neural network models, respectively. 

More recently, \citet{pu-etal-2023-incorporating} proposed an approach that incorporates the uncertainty of RST structures into the attention mechanisms of summarization models and achieved SOTA results on multiple datasets. However, all the above approaches require full fine-tuning of NLG models, which is very expensive. As the model parameters increase, this issue will be further amplified. Incorporating RST into PEFT might potentially lower the barrier to fine-tuning by structuring the learning process around the inherent rhetorical patterns in the data. 

\subsection{Document Summarization with LoRA}

LoRA, presented by \citet{hu2022lora}, is a low-rank approximation strategy that reduces the number of trainable parameters by freezing the pre-trained model weights and injecting trainable rank decomposition matrices into each layer of the transformer architecture. The initial research also demonstrated through summarization tasks that applying LoRA on the GPT-3 model \cite{NEURIPS2020_1457c0d6} with less than 1\% of the parameters could even outperform FFT. Expanding on this, studies by \citet{dettmers2023qlora}, \citet{xu2023qalora}, and \citet{li2023loftq} enhanced generalization ability in downstream summarization tasks by quantifying LoRA matrices and adopting mixed-precision techniques. Furthermore, \citet{zhu2023parameter} combined LoRA with layer pruning, achieving notable improvements in specialized applications like medical report summarization. Recently, \citet{liao-etal-2023-parameter} validated the feasibility of using task-neutral sparse masks to improve the performance in text summarization with LoRA.

In a similar work, \citet{ghazvininejad-etal-2022-discourse} integrated hierarchical document structure (i.e. blocking structure) into prefix-tuning to simulate the high-level discourse relation and achieved improvements in the task of text generation. However, there is still an unexplored potential in explicitly integrating fine-grained RST structures into the summarization process with PEFT methods, since comprehending the coherence of discourse elements could positively impact the quality of generated summaries \cite{li-etal-2016-role, liu-chen-2019-exploiting, huang-kurohashi-2021-extractive}, particularly in the context of summarizing long documents \cite{cohan-etal-2018-discourse, xu-etal-2020-discourse, li-etal-2020-composing, gabriel-etal-2021-discourse, dong-etal-2021-discourse, balachandran-etal-2022-correcting, pu-etal-2023-incorporating}.

\section{Proposed Approach}
A limitation observed with LoRA and other PEFT methods during the fine-tuning phase is that, while their primary function is to act as a low-rank approximation of the weight matrices in LLMs, they do not adequately capture textual context relations \cite{he2022towards, ghazvininejad-etal-2022-discourse, wan-etal-2023-pip, zhang-etal-2023-towards-adaptive}. One of the reasons is that LoRA is not driven or guided by discourse knowledge during the training phase, because this part of knowledge is not explicitly present in the input data \cite{ustun-cooper-stickland-2022-parameter, zhao-etal-2023-infusing}. In addition, the matrices obtained by low-rank approximation strategies may have more difficulty in capturing complex textual discourse relations due to the smaller semantic space that can be expressed when compared to LLMs' weight matrices \cite{wan-etal-2023-pip, zhang-etal-2023-towards-adaptive, tomanek-etal-2021-residual}. Hence, we propose a method that directly and explicitly incorporates discourse architecture into LoRA. This approach allows LoRA’s adaptation mechanism for downstream tasks to discern intricate text relations through soft guidance, which can leverage contextual discourse connections and steer the learning trajectory toward a discourse-informed summarization process.

\begin{figure}[ht]
  \centering
  \includegraphics[width=0.415\textwidth]{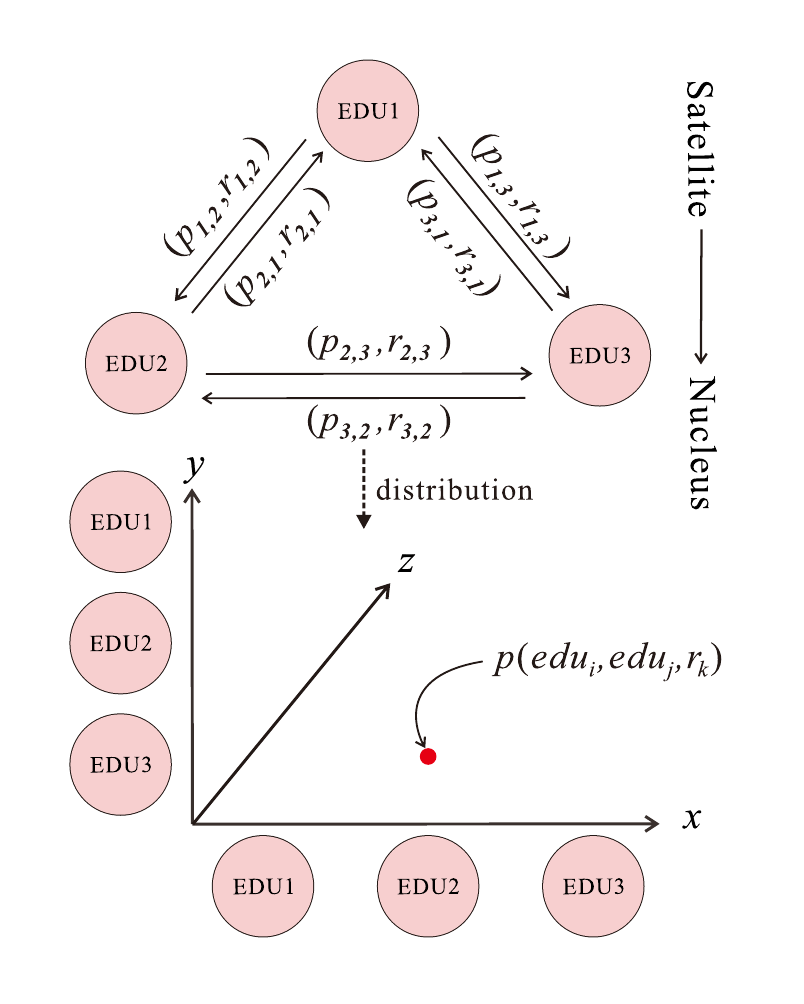}
  \caption{RST distribution}
  \label{fig:RST_distribution}
\end{figure}

\subsection{RST Distribution}

Our approach builds upon prior practice \cite{chen-etal-2017-improved, xiao-etal-2020-really, bugliarello-okazaki-2020-enhancing, pu-simaan-2022-passing, pu-etal-2023-incorporating, zhao-etal-2023-infusing} of integrating linguistic structures (such as syntactic structure, discourse structure, etc.) into neural NLG models. To infuse discourse structure into LoRA, we begin by converting RST structures, generated by an RST parser\footnote{\url{https://github.com/seq-to-mind/DMRST_Parser}} into a compact matrix representation. Figure \ref{fig:RST_distribution} exemplifies how to transmute the full potential RST structures (n-best RST forests) into a three-dimensional discourse matrix \cite{pu-etal-2023-incorporating}. In this matrix, the $x$ and $y$ axes correspond to the elementary discourse units (EDUs) within the source document, while the $z$-axis denotes the discourse relation label\footnote{Appendix \ref{sec:appendix0} details the grouping of discourse relations.}. Each point of the matrix indicates the probability value \( p(edu_i, edu_j, r_k) \in [0, 1] \subseteq \mathbb{R} \) that \( edu_i \) is the nucleus of \( edu_j \) with discourse relation \( r_k \). It should be noted that $\forall i = j$, \( p(edu_i, edu_j, r_k) = 0 \), since no unit is self-dependent. Next, we average and merge the $y$-axis of the matrix, and the merged value $c(edu_i,\overline{edu_j},r_k)$ is called the importance index of $edu_i$ with relation $r_k$. The $RST$ distribution is then obtained by combining all $c(edu_i,\overline{edu_j},r_k)$. Based on this, we propose four fine-grained RST matrix distributions:

\begin{itemize}[leftmargin=8pt,itemsep=1pt,topsep=1pt,parsep=1pt]
    \item $RST^{b}_{wo}$: A binary, label-agnostic representation collapsing probabilities into a simple 1-or-0 regarding discourse connections.
    \item $RST^{b}_{w}$: An extension of the binary distribution that includes relation labels, enriching the binary decisions with relational types.
    \item $RST^{p}_{wo}$: A probabilistic representation that omits labels, focusing instead on the probabilities to express uncertainty in discourse connections.
    \item $RST^{p}_{w}$: The most granular representation, retaining both types of discourse relations and their probabilistic weights for a full-fledged representation of discourse nuances.
\end{itemize}

The inclusion of the relation label is contingent on whether we perform average-and-merge along the relation dimension ($z$-axis). Whether the approach is binary or based on uncertainty hinges on whether we replace the probability value with 1 or 0. In the binary cases, probabilities equal to or above 0.5 are replaced with 1, else with 0. Previous researchers (such as \citet{xu-etal-2020-discourse} and \citet{dong-etal-2021-discourse}) considered the 1-best tree, representing binary relations outputted from parsers into summarization models (also the case of our first two variants). The latter two variants utilize the parser's output probabilities as confidence indicators for discourse connections \cite{pu-etal-2023-incorporating}. This probabilistic approach softens the impact of potential errors or ambiguities in the parser's output, blending its uncertainty into the model. 

\subsection{RST-Aware Injection}

\begin{figure*}[t]
\centering
\includegraphics[width=0.83\textwidth]{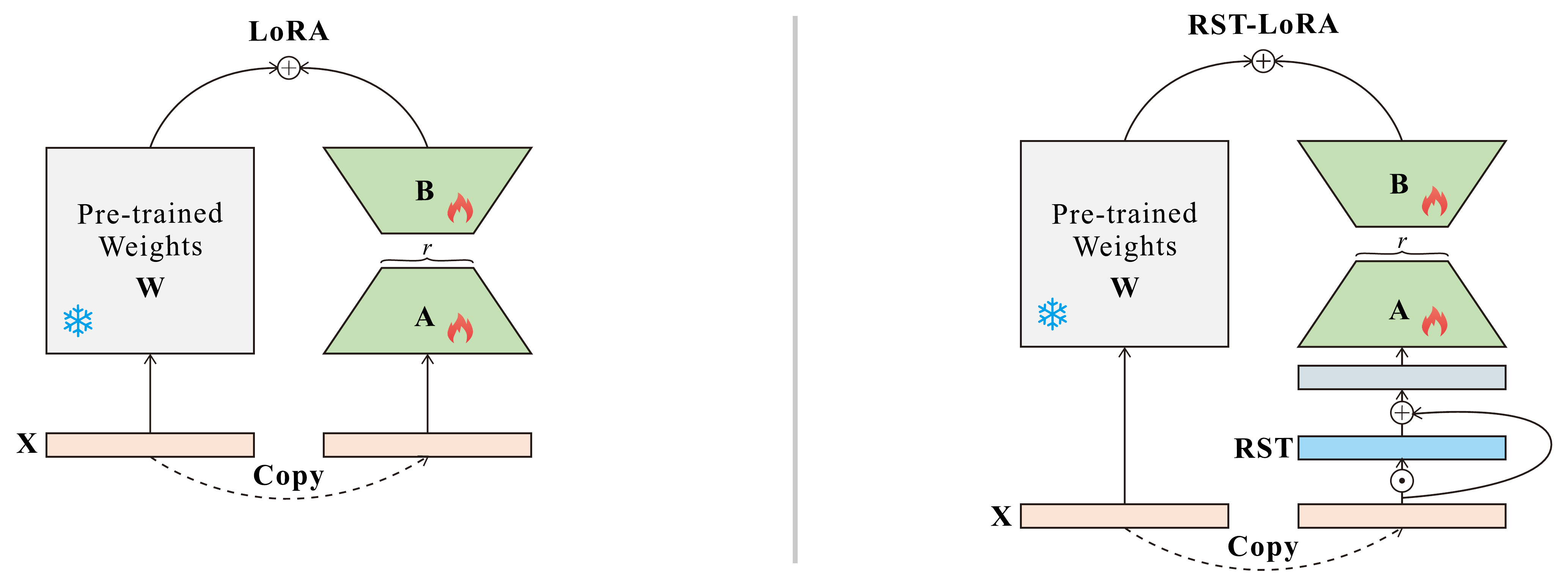}
\caption{Model architecture: The diagram illustrates the integration of the RST matrix into the LoRA model. The left side is the original LoRA, while the right side depicts our proposed method RST-LoRA.} 
\label{fig:model}
\end{figure*}

In the process of vanilla LoRA fine-tuning, let $W^{fine-tuned}_{A \times B}$ denote the fine-tuned LLM's parameters, and $W^{pre-trained}_{A \times B}$ represent the parameters before fine-tuning. The change in parameters is represented by $\Delta W_{A \times B}$, where $A$ and $B$ correspond to the dimensions of the parameter matrix:

\[ W^{fine-tuned}_{A \times B}  =  W^{pre-trained}_{A \times B}  +  \Delta W_{A \times B} \]

In other words, the parameters after fine-tuning can be obtained by adding a matrix representing the variation to the parameters of the original, pre-fine-tuned model.

\[    \Delta W_{A \times B}  \simeq  \Phi [(W^{down}_{A \times r} W^{up}_{r \times B})]\]
\[
    r \ll min(A, B)
\]

The objective of the LoRA strategy aims to learn the mapping method $\Phi$ that can provide an approximation of the matrix representing parameter variations \cite{hu2022lora}. Typically, the rank value $r$ is considerably smaller than both $A$ and $B$, so that the total number of parameters of $W^{down}_{A \times r}$ and $W^{up}_{r \times B}$ is significantly smaller than $W_{A \times B}$. For a given input document $X$ to the linear projection in the model's hidden layer, LoRA modifies the projection output (hidden representation) $h$ as follows:

 \[   h \leftarrow h + X (W^{down}_{A \times r} W^{up}_{r \times B}) \]

In its current form, LoRA treats both satellite and nucleus EDUs in documents equally and only recognizes their difference during the back-propagation process. This issue is also noted in the analyses by \citet{ghazvininejad-etal-2022-discourse, zhao-etal-2023-infusing}, who also discovered that PEFT faces challenges in understanding the complex relations between sentences and the differences in importance level between text segments during its learning process. Therefore, we soft-guide the learning process by injecting the RST structure (i.e., the matrix presentation mentioned above) into the text embedding matrix of LoRA, as shown in Figure \ref{fig:model}. Specifically:

\[h \leftarrow h + [(X\odot (1+\gamma)) (W^{down}_{A \times r} W^{up}_{r \times B})] \]

Here, $\gamma$ denotes the weight coefficient matrix, or more precisely, the RST distribution matrix. The operation $\odot$ signifies element-wise multiplication, and the motivation behind employing element-wise multiplication is that it can significantly amplify the impact of probability values on the input $X$ matrix, creating an RST-injected matrix with greater distributional variance, in contrast, element-wise addition would exert a lesser impact on $X$. It should be noted that RST parser operates at the EDU level, meaning that sub-word units within the same EDU share the same multiplication factor, embedding the same probability value across the entire EDU into $X$. The estimates of learned parameters $W^{down}_{A \times r}$ and $W^{up}_{r \times B}$ are adjusted to match the utility of discourse knowledge for the ultimate summarization purpose. Each element of $\gamma$ is constrained to be non-negative. The operation of $1+\gamma$ functions as a residual connection, allowing discourse knowledge to exert a subtle influence on the adjustment of the low-rank weight matrix. If we set all elements of $\gamma$ to a uniform value $\delta$, including zero, the adjustment to the low-rank matrices would revert to the conventional LoRA approach.

\section{Experiments and Analysis}

\subsection{Experimental Settings}

\paragraph{Datasets}

Our experiments are conducted on three recent long document summarization datasets: Multi-LexSum \citep[ML,][]{shen2022multi}, eLife \cite{goldsack-etal-2022-making} and BookSum Chapter \citep[BC,][]{kryscinski-etal-2022-booksum}. These datasets are sourced from the fields of legal documents, scientific papers, and books, respectively. We select these datasets because they exhibit a high degree of heterogeneity, and we want to test whether our proposed approach could maintain sufficient generalization performance across different data domains. Statistics of these datasets are provided in Appendix \ref{sec:appendix0_1}. 

\paragraph{Parser}
For automatic parsing of source documents, we employ DMRST parser \cite{liu-etal-2020-multilingual-neural, liu-etal-2021-dmrst} which enables us to extract probabilities or uncertainties of discourse relations and type labels from its final logits layer.

\paragraph{Automatic Metrics}

Aligning with previous work for evaluating summarization systems \cite{narayan-etal-2018-dont, liu-etal-2023-binary, blinova-etal-2023-simsum}, we use F1 scores of Rouge-1 (R1), Rouge-2 (R2), Rouge-L (RL), and Rouge-Lsum (RLsum) \cite{lin-2004-rouge}, BERTScore \cite{zhang2019bertscore}, METEOR \cite{banerjee-lavie-2005-meteor}, sacreBLEU \cite{post-2018-call} and NIST \cite{lin-hovy-2003-automatic} for model’s performance evaluation. A description of these metrics can be found in Appendix \ref{sec:appendix0_2}.

\paragraph{Training \& Inference}
We operate Longformer \cite{beltagy2020longformer} and Vicuna13B-16k \cite{zheng2023judging} as our baseline backbone models. Longformer is a state-of-the-art, open-source model optimized for handling long documents under Seq2Seq architecture. Meanwhile, Vicuna is another SOTA model based on GPT architecture. Our objective in using these models is to demonstrate the generalizability of our strategy across different architectural frameworks. We also include GPT-4 \cite{OpenAI2023GPT4TR} as one of our comparative models. It should be noted that for GPT-4, we use both zero-shot learning (ZS) and in-context learning (ICL) with demonstrations from two randomly selected samples from the training datasets\footnote{Prompts can be found in Appendix \ref{sec:appendix1_0} and \ref{sec:appendix1_1}.}. Besides, we compare our results with both the original full parameter fine-tuning (FFT) and the vanilla LoRA fine-tuning. All open-source models, including the baseline, proposed, and ablation models, adhere to identical hyperparameter settings. These settings are elaborated in Appendix \ref{sec:appendix1}.

\subsection{Experimental Results}

\paragraph{General Results}

The differences in performance of different RST variants are shown in Table \ref{tab:RST-variants}. Among our proposed RST-injected variants, models integrating discourse relation labels generally outperformed those without this integration. Similarly, models considering the uncertainty in discourse relations fare better than those disregarding it. This suggests that integrating parser uncertainty and coherence labels into the model improves the robustness of the model against potential misinformation to a certain extent when compared to the parser's 1-best binary decisions.

\begin{table}[ht]
\centering
\scalebox{0.68}{
\tabcolsep=4pt
\begin{threeparttable}
\begin{tabular}{c l c c c c}
\toprule
Data & Model  & R1$_{f1}$$\uparrow$ & R2$_{f1}$$\uparrow$ & RL$_{f1}$$\uparrow$ & RLsum$_{f1}$$\uparrow$ \\
\hline

\multirow{8}*{\rotatebox[origin=c]{90}{Multi-LexSum}} 
~ & Longformer\mbox{$_{RST^b_{wo}-LoRA}$} & 45.82 & 21.32 & 23.81 & 43.40 \\
~ & Longformer\mbox{$_{RST^b_{w}-LoRA}$} & 46.02 & 21.34 & 23.87 & 43.39 \\
~ & Longformer\mbox{$_{RST^p_{wo}-LoRA}$} & 46.21 & 21.54 & 24.09 & 43.37  \\
~ & Longformer\mbox{$_{RST^p_{w}-LoRA}$} & 46.33 & 21.86 & 24.11 & 43.58 \\
\hhline{~|-----}
~ & Vicuna\mbox{$_{RST^b_{wo}-LoRA}$} & 46.32 & 21.64 & 24.22 & 43.32 \\
~ & Vicuna\mbox{$_{RST^b_{w}-LoRA}$} & 47.33 & 22.70 & 24.25 & 43.31 \\
~ & Vicuna\mbox{$_{RST^p_{wo}-LoRA}$} & 47.39 & 22.79 & 24.35 & 43.33 \\
~ & Vicuna\mbox{$_{RST^p_{w}-LoRA}$} & 47.45 & 23.19 & 24.39 & 44.02 \\
\midrule
\midrule
\multirow{8}*{\rotatebox[origin=c]{90}{eLife}}
~ & Longformer\mbox{$_{RST^b_{wo}-LoRA}$} & 49.34 & 14.24 & 21.34 & 46.74 \\
~ & Longformer\mbox{$_{RST^b_{w}-LoRA}$} & 49.41 & 14.39 & 21.29 & 46.79 \\
~ & Longformer\mbox{$_{RST^p_{wo}-LoRA}$} & 49.87 & 14.49 & 21.83 & 47.15 \\
~ & Longformer\mbox{$_{RST^p_{w}-LoRA}$} & 49.89 & 14.68 & 22.11 & 47.64 \\
\hhline{~|-----}
~ & Vicuna\mbox{$_{RST^b_{wo}-LoRA}$} & 48.73 & 14.68 & 21.89 & 47.11 \\
~ & Vicuna\mbox{$_{RST^b_{w}-LoRA}$} & 49.72 & 14.72 & 22.03 & 47.02 \\
~ & Vicuna\mbox{$_{RST^p_{wo}-LoRA}$} & 49.87 & 14.79 & 22.21 & 48.10 \\
~ & Vicuna\mbox{$_{RST^p_{w}-LoRA}$} & 49.92 & 14.92 & 22.41 & 48.21 \\
\midrule
\midrule
\multirow{8}*{\rotatebox[origin=c]{90}{BookSum Chapter}}
~ & Longformer\mbox{$_{RST^b_{wo}-LoRA}$} & 34.70 & 10.22 & 20.39 & 34.21  \\
~ & Longformer\mbox{$_{RST^b_{w}-LoRA}$} & 34.72 & 10.19 & 20.41 & 34.87 \\
~ & Longformer\mbox{$_{RST^p_{wo}-LoRA}$} & 35.29 & 11.38 & 21.62 & 35.11 \\
~ & Longformer\mbox{$_{RST^p_{w}-LoRA}$} & 35.40 & 11.76 & 21.88 & 35.27 \\
\hhline{~|-----}
& Vicuna\mbox{$_{RST^b_{wo}-LoRA}$} & 37.28 & 12.35 & 22.13 & 38.33 \\
~ & Vicuna\mbox{$_{RST^b_{w}-LoRA}$} & 37.41 & 12.66 & 22.51 & 38.40 \\
~ & Vicuna\mbox{$_{RST^p_{wo}-LoRA}$} & 37.87 & 13.10 & 22.77 & 39.69 \\
~ & Vicuna\mbox{$_{RST^p_{w}-LoRA}$} & 37.92 & 13.24 & 22.93 & 40.31 \\
\bottomrule
\end{tabular}
\end{threeparttable}
}
\caption{Performance of different RST variants}
\label{tab:RST-variants}
\end{table}

\begin{table*}[t]
\centering
\scalebox{0.65}{
\tabcolsep=4pt
\begin{threeparttable}
\begin{tabular}{c l c c c c c c c c c}
\toprule
Dataset & Model &  \# Trainable
Parameters & R1$_{f1}$$\uparrow$ & R2$_{f1}$$\uparrow$ & RL$_{f1}$$\uparrow$ & RLsum$_{f1}$$\uparrow$ & BERTscore$_{f1}$$\uparrow$ & Meteor$\uparrow$ & sacreBLEU$\uparrow$ & NIST$\uparrow$ \\
\hline
\multirow{10}*{\rotatebox[origin=c]{90}{Multi-LexSum}} 
& Longformer$_{FFT}$ & 0.44B & 45.81 & 21.32 & 23.71 & 43.25 & 87.21 & 33.30 & 12.06 & 2.23\\
~ & Longformer$_{LoRA}$ & 1.13M & 45.78 & 21.30 & 23.65 & 43.12 & 87.31 & 33.31 & 12.00 & 2.28 \\
~ & Longformer\mbox{$_{RST^p_{w}-LoRA}$} & 1.13M & 46.33$^\dag$$^\ddag$ & 21.86$^\dag$$^\ddag$ & 24.11$^\dag$$^\ddag$ & 43.58$^\dag$$^\ddag$ & 92.01$^\dag$$^\ddag$ & 34.55$^\dag$$^\ddag$ & 13.11$^\dag$$^\ddag$ & 3.21$^\dag$$^\ddag$ \\
~ & Vicuna$_{FFT}$ & 13B & 46.40 & 21.88 & 24.15 & 43.28 & 90.02 & 33.19 & 13.56 & 3.32 \\
~ & Vicuna$_{LoRA}$ & 6M & 46.32 & 21.76 & 24.09 & 43.14 & 89.45 & 33.22 & 13.44 & 3.31 \\
~ & Vicuna\mbox{$_{RST^p_{w}-LoRA}$} & 6M & 47.45$^\ddag$ & 23.19$^\dag$$^\ddag$ & \cellcolor{Peach}\textbf{24.39}$^\dag$$^\ddag$ & \cellcolor{Peach}\textbf{44.02} $^\dag$$^\ddag$ & \cellcolor{Peach}\textbf{93.89}$^\dag$$^\ddag$ & \cellcolor{Peach}\textbf{35.31}$^\dag$$^\ddag$ & \cellcolor{Peach}\textbf{14.02}$^\dag$$^\ddag$ & \cellcolor{Peach}\textbf{4.11}$^\dag$$^\ddag$ \\
~ & GPT-4$_{ZS}$ & - & 38.74 & 13.39 & 18.26 & 37.67 & 60.91 & 24.24 & 7.43 & 1.55 \\
~ & GPT-4$_{ICL}$ & - & 42.14 & 15.27 & 20.37 & 40.12 & 71.32 & 28.14 & 10.22 & 1.90 \\
~ & \citet{pu-etal-2023-incorporating} & - & 46.42 & 22.89 & - & 43.98 & 86.70 & 33.94 & - & - \\
~ & \citet{shen2022multi} & - & \cellcolor{Peach}\textbf{53.73} & \cellcolor{Peach}\textbf{27.32} & - & 30.89 & 42.01 & - & - & - \\
\midrule
\midrule
\multirow{10}*{\rotatebox[origin=c]{90}{eLife}}
& Longformer$_{FFT}$ & 0.44B & 47.59 & 13.58 & 20.75 & 45.25 & 85.50 & 28.21 & 6.86 & 2.90 \\
~ & Longformer$_{LoRA}$ & 1.13M & 48.31 & 13.69 & 21.10 & 45.80 & 85.63 & 28.18 & 7.05 & 3.12 \\
~ & Longformer\mbox{$_{RST^p_{w}-LoRA}$} & 1.13M & 49.89$^\dag$$^\ddag$ & 14.68$^\dag$$^\ddag$ & 22.11$^\dag$$^\ddag$ & 47.64$^\dag$$^\ddag$ & 87.64$^\dag$$^\ddag$ & 31.23$^\dag$$^\ddag$ & 7.78$^\dag$$^\ddag$ & \cellcolor{Peach}\textbf{3.79}$^\dag$$^\ddag$ \\
~ & Vicuna$_{FFT}$ & 13B & 48.32 & 14.06 & 21.31 & 45.57 & 85.71 & 30.28 & 7.00 & 2.91 \\
~ & Vicuna$_{LoRA}$ & 6M & 48.41 & 14.32 & 21.40 & 46.01 & 86.06 & 31.00 & 6.62 & 2.88 \\
~ & Vicuna\mbox{$_{RST^p_{w}-LoRA}$} & 6M & \cellcolor{Peach}\textbf{49.92}$^\dag$$^\ddag$ & \cellcolor{Peach}\textbf{14.92}$^\dag$$^\ddag$ & \cellcolor{Peach}\textbf{22.41}$^\dag$$^\ddag$ & \cellcolor{Peach}\textbf{48.21}$^\dag$$^\ddag$ & \cellcolor{Peach}\textbf{87.81}$^\dag$$^\ddag$ & \cellcolor{Peach}\textbf{33.22}$^\dag$$^\ddag$ & \cellcolor{Peach}\textbf{8.15}$^\dag$$^\ddag$ & 3.42$^\dag$$^\ddag$\\
~ & GPT-4$_{ZS}$ & - & 42.73 & 9.05 & 17.93 & 40.15 & 61.21 & 25.13 & 3.47 & 2.32 \\
~ & GPT-4$_{ICL}$ & - & 44.62 & 11.35 & 20.03 & 44.09 & 73.23 & 27.36 & 5.66 & 2.45 \\
~ & \citet{pu-etal-2023-incorporating} & - & 48.70 & 14.84 & - & 46.13 & 84.70 & 29.53 & - & - \\
\midrule
\midrule
\multirow{11}*{\rotatebox[origin=c]{90}{BookSum Chapter}}
~ & Longformer$_{FFT}$ & 0.44B & 34.68 & 10.02 & 20.35 & 33.71 & 81.02 & 27.30 & 3.32 & 1.62\\
~ & Longformer$_{LoRA}$ & 1.13M & 34.63 & 9.96 & 20.22 & 33.79 & 81.33 & 27.32 & 3.55 & 1.86 \\
~ & Longformer\mbox{$_{RST^p_{w}-LoRA}$} & 1.13M & 35.40$^\dag$$^\ddag$ & 11.76$^\dag$$^\ddag$ & 21.88$^\dag$$^\ddag$ & 35.27$^\dag$$^\ddag$ & 83.99$^\dag$$^\ddag$ & 29.03$^\dag$$^\ddag$ & \cellcolor{Peach}\textbf{5.94}$^\dag$$^\ddag$ & 2.02$^\dag$$^\ddag$ \\
~ & Vicuna$_{FFT}$ & 13B & 37.21 & 12.38 & 22.07 & 38.21 & 82.31 & 28.01 & 3.45 & 1.70 \\
~ & Vicuna$_{LoRA}$ & 6M & 37.30 & 12.26 & 21.84 & 38.23 & 82.23 & 27.83 & 3.34 & 1.68 \\
~ & Vicuna\mbox{$_{RST^p_{w}-LoRA}$} & 6M & 37.92$^\dag$$^\ddag$ & \cellcolor{Peach}\textbf{13.24}$^\dag$$^\ddag$ & \cellcolor{Peach}\textbf{22.93}$^\dag$$^\ddag$ & \cellcolor{Peach}\textbf{40.31}$^\dag$$^\ddag$ & 84.12$^\dag$$^\ddag$ & \cellcolor{Peach}\textbf{29.22}$^\dag$$^\ddag$ & 5.48$^\dag$$^\ddag$ & \cellcolor{Peach}\textbf{2.32}$^\dag$$^\ddag$ \\
~ & GPT-4$_{ZS}$ & - & 35.25 & 7.46 & 17.52 & 34.23 & 58.56 & 26.50 & 3.36 & 1.54 \\
~ & GPT-4$_{ICL}$ & - & 37.42 & 10.06 & 19.49 & 36.11 & 79.56 & 27.56 & 3.52 & 1.72 \\
~ & \citet{pu-etal-2023-incorporating} & - & 34.02 & 10.28 & - & 32.87 & \cellcolor{Peach}\textbf{85.30} & 27.47 & - & - \\
~ & \citet{cao2023awesome} & - & 41.11 & 10.63 & - & 40.20 & - & - & - & - \\
~ & \citet{scire-etal-2023-echoes} & - & \cellcolor{Peach}\textbf{42.13} & 10.53 &  16.75 & - & - & - & - & - \\
\bottomrule
\end{tabular}
\end{threeparttable}
}
\caption{Model performance. The bold numbers represent the best results concerning the given test set. $^\dag$ and $^\ddag$ indicate statistical significance (p$<$0.05) of our final model (\mbox{$\mathrm{RST^p_{w}}$-LoRA}) against the FFT and LoRA model via paired t-test based on the same backbone respectively. $FFT$ for full fine-tuning, $ZS$ for zero-shot learning and $ICL$ for in-context learning. Each result of the SOTA models is directly replicated from their original papers.}
\label{tab:model_performance}
\end{table*}

Table \ref{tab:model_performance} shows the performance differences between our final strategy (the best RST variant) and other comparative models. Specifically, GPT-4 exhibits the poorest overall performance, attributable to a lack of parameter tuning. The performance of the models based on Vicuna as backbone is overall better than the models based on Longformer due to the larger number of parameters. Regarding parameter-efficient settings, vanilla LoRA's performance is marginally lower than FFT across most datasets, except eLife. However, LoRA achieves comparable results to FFT while only requiring adjustments of 0.25\% of parameters for Longformer and 0.05\% for Vicuna, highlighting LoRA's efficiency.

We also observe consistent performance improvements in LoRA when integrating RST structure into its training process without increasing the number of fine-tunable parameters, and in most cases even exceeds the FFT model. Our final model \mbox{$\mathrm{RST^p_{w}}$-LoRA}, integrates both discourse relation types and uncertainty into LoRA’s training, achieving the best experimental outcomes. It also defeats SOTA models (fully fine-tuned with complicated strategies) on some metrics, including the current most advanced model \cite{pu-etal-2023-incorporating} that incorporates RST structure to improve summarization performance.

\paragraph{Ablation Results}

To further assess the impact of the RST matrix on model performance, we specify three additional control conditions:

\begin{itemize}[leftmargin=8pt,itemsep=1pt,topsep=1pt,parsep=1pt]
    \item RST$_{Even}$: In the RST matrix, we set values to 1 at even positions and 0 at odd positions.
    \item RST$_{Odd}$: We assign values of 1 at odd positions and 0 at even positions in the RST matrix.
    \item RST$_{Random}$: We assign random values $\in [0, 1] \subseteq \mathbb{R}$ to the RST matrix without considering the probability of discourse relations.
\end{itemize}

\begin{table}[htbp]
\centering
\tabcolsep=4pt
\scalebox{0.78}{
\begin{threeparttable}
\begin{tabular}{c l c c c c}
\toprule
Dataset & Model & R1$_{f1}$$\uparrow$ & R2$_{f1}$$\uparrow$ & RL$_{f1}$$\uparrow$ & RLsum$_{f1}$$\uparrow$\\

\hline
\multirow{3}*{ML}
& RST$_{Even}$ & 46.21 & 21.39 & 23.66 & 42.55\\
~ & RST$_{Odd}$ & 46.26 & 21.37 & 23.82 & 42.90\\
~ & RST$_{Random}$ & 46.30 & 21.73 & 24.07 & 43.10\\
\hdashline
\multirow{3}*{eLife} 
& RST$_{Even}$ & 47.10 & 14.28 & 20.86 & 45.33\\
~ & RST$_{Odd}$ & 47.04 & 14.20 & 20.98 & 45.31\\
~ & RST$_{Random}$ & 47.32 & 14.29 & 21.36 & 45.71\\
\hdashline
\multirow{3}*{BC} 
& RST$_{Even}$ & 37.09 & 12.20 & 21.75 & 38.06\\
~ & RST$_{Odd}$ & 37.01 & 12.18 & 21.72 & 38.10\\
~ & RST$_{Random}$ & 37.27 & 12.23 & 21.80 & 38.19\\
\bottomrule
\end{tabular}
\end{threeparttable}
}
\caption{F1 scores for ablation study}
\label{tab:ablation_study}
\end{table}

In ablation experiments, we use Vicuna as backbone for testing. The motivation behind setting these three ablation conditions is to simulate the extreme scenario where the RST parser completely fails to deliver valuable discourse information. Table \ref{tab:ablation_study} indicates that different ablation integration strategies not only fail to enhance the model's performance but even detract from it. Experiments by introducing random noise exhibit that these arbitrary values reduce the model's performance to a level marginally lower than the original LoRA. Furthermore, this also implies that when the RST parser fails to provide meaningful knowledge (as in the case of random noise), the impact of noise on the performance of the model is limited.

\subsection{Analysis}

\begin{figure}[h]
  \centering
  \includegraphics[width=0.49\textwidth]{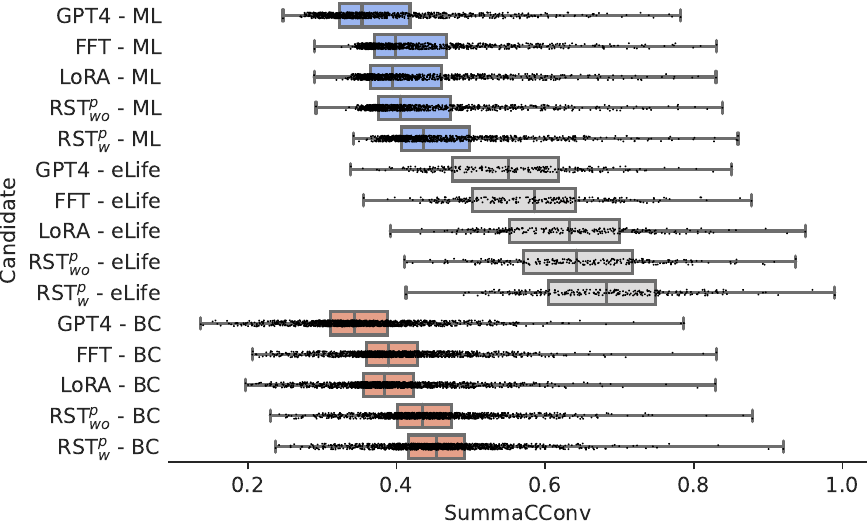}
  \caption{Factual consistency analysis}
  \label{fig:SummC}
\end{figure}

\paragraph{Hallucination Checking}
We delve deeper into the level of factual consistency of the generated summaries, which we test using the SummaC method \cite{laban-etal-2022-summac}. The score of SummaC ranges from 0 to 1, and the higher the score, the better the consistency. The results of the assessment using Vicuna as backbone are depicted in Figure \ref{fig:SummC}. We observe that GPT-4 exhibits the weakest factual consistency, while the original LoRA also shows a comparatively lower level of factual accuracy than FFT. However, explicitly incorporating RST structure into LoRA mitigates the issue of hallucinations/inaccuracies in generated summaries, achieving better results than FFT model.

\paragraph{Impact of Different Rank $r$}

Figure \ref{fig:diff_r_ML} and Figures \ref{fig:diff_r_elife}, \ref{fig:diff_r_BC} in Appendix \ref{sec:appendix2} illustrate the impact of different ranks on model performance (Vicuna backbone). Across different datasets, the RST-aware model consistently outperforms the original LoRA at various ranks and achieves similar performance as the FFT model at lower ranks. Furthermore, a larger rank $r$ will help to improve the performance of the model, which is also aligned with the findings of \citet{he2022towards, zhang2023adaptive}. However, a higher rank correlates with an increased number of parameters requiring adjustment. Importantly, \( r \) = 8 is a trade-off point between performance gain and computational cost, when $r$ continues to increase, the gain rate of performance improvement begins to slow down.

\begin{figure}[ht]
  \centering
  \includegraphics[width=0.47\textwidth]{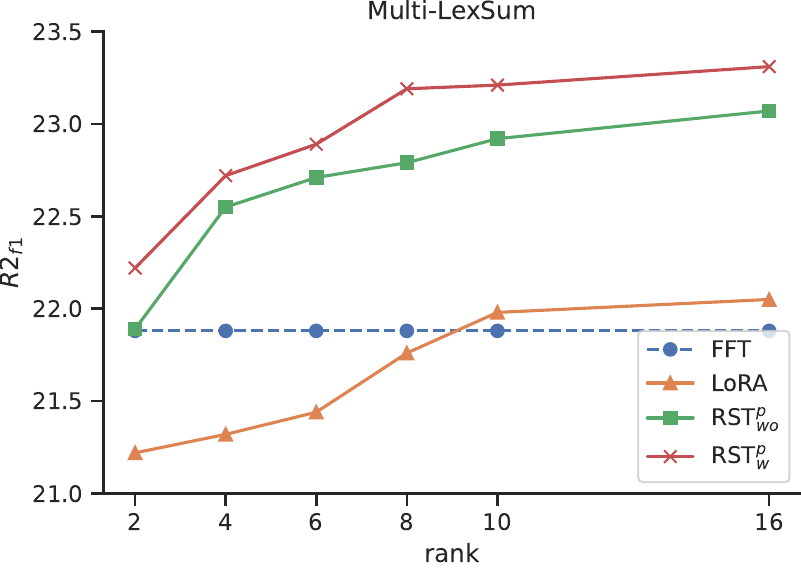}
  \caption{Impact of different $r$ on ML dataset}
  \label{fig:diff_r_ML}
\end{figure}

\paragraph{Impact of Parser Capability}

To rigorously evaluate the parser's impact on our method, we conduct an experiment that involves intentionally altering the RST parser's output. This is designed to simulate varying levels of parser performance instability, thereby allowing us to observe its influence on our model's efficacy. Specifically, we introduce random masking to the parser's output at incremental thresholds of 10\%, 20\%, 40\%, and 80\%, assigning random values within the range of 0 to 1 to portions of the RST matrix. Table \ref{tab:parser-impact} presents the findings from this experiment, with Vicuna serving as backbone for \mbox{$\mathrm{RST^p_{w}}$-LoRA} model on the Multi-LexSum dataset.
 
\begin{table}[ht]
\tabcolsep=6pt
\centering
\scalebox{0.88}{
\begin{threeparttable}
\begin{tabular}{ccccc}
\toprule
Model & R1$_{f1}$$\uparrow$ & R2$_{f1}$$\uparrow$ & RL$_{f1}$$\uparrow$ & RLsum$_{f1}$$\uparrow$\\
\midrule
RST\_10\% & 47.33 & 23.01 & 24.33 & 43.45 \\
RST\_20\% & 47.09 & 22.78 & 24.23 & 43.37 \\
RST\_40\% & 46.52 & 21.76 & 24.13 & 43.20 \\
RST\_80\% & 46.32 & 21.75 & 24.06 & 43.15 \\
\bottomrule
\end{tabular}
\end{threeparttable}
}
\caption{Impact of random masking on the parser}
\label{tab:parser-impact}
\end{table}

These results illustrate the direct correlation between the RST parser's performance and the performance of our model's output. Notably, even under conditions of compromised parser performance (with up to 20\% of the information being randomly masked), our model still demonstrates a good capacity to enhance summary generation quality by leveraging the learned discourse structure knowledge. However, it is observed that when the level of induced noise surpassed 40\%, the negative impact became pronounced, relegating the model's performance to levels akin to that of the original vanilla LoRA.

\paragraph{Human Evaluation}

To better analyze the quality of the summaries generated by the models, we randomly select 10 instances from the BookSum dataset and conduct a human evaluation. The evaluators we have recruited are graduate and doctoral candidates with specializations in Computer Science or Computational Linguistics, each possessing advanced proficiency in English. They receive compensation at the University's established hourly rate. Evaluators are asked to read the corresponding original document, as well as five candidate summaries (from FFT, LoRA, and \mbox{$\mathrm{RST^p_{w}}$-LoRA} with Vicuna backbone, GPT-4 and human). The human evaluators are blind to the condition, i.e.~they do not know which summary comes from which system (or human author). Each sample is independently evaluated by three distinct human raters (thus 150 evaluation samples in total). Evaluators should rate the candidate summaries on a scale of 1 to 5 for relevance (R), informativeness (I), conciseness (C), and faithfulness (F), with a higher score indicating better quality. They also need to give an overall ranking of the five summaries. The detailed guidelines for human evaluation are available in Appendix \ref{sec:appendix3}. The results, presented in Table \ref{tab:human_evaluation}, show the average values for each metric, as well as the proportions of times each model's output is considered the best or worst among the candidates. The scores of Fleiss' Kappa coefficient for R, I, C, and F are 0.812, 0.705, 0.683, and 0.688, respectively, with an average score of 0.722, indicating substantial agreement.

\begin{table}[t]
\tabcolsep=2pt
\centering
\scalebox{0.85}{
\begin{threeparttable}
\begin{tabular}{lccccc}
\toprule
Candidate & R & I & C & F & Best $\mid$ Worst\\
\hline
Human & 4.70 & 4.83 & 4.53 & 4.67 & 83.3\% $\mid$ 0.0\% \\
\hdashline
GPT-4$_{ICL}$ & 3.76 & 2.27 & 3.25 & 2.33 & 0.0\% $\mid$ 56.7\% \\
Vicuna$_{LoRA}$ & 4.03 & 2.37 & 3.20 & 2.50 & 0.0\% $\mid$ 20.0\% \\
Vicuna$_{FFT}$ & 4.27 & 2.57 & 3.67 & 2.77 & 6.67\% $\mid$ 13.3\% \\
Vicuna\mbox{$_{RST^p_{w}-LoRA}$} & 4.53 & 3.90 & 4.03 & 3.17 & 13.3\% $\mid$ 10.0\% \\
\bottomrule
\end{tabular}
\end{threeparttable}
}
\caption{Human evaluation results}
\label{tab:human_evaluation}
\end{table}

From Table \ref{tab:human_evaluation}, it is evident that human-generated summaries surpass all neural summarization models in terms of quality. Among the four neural models, GPT-4 shows the least performance, with LoRA coming in second, having a 20\% probability of being rated as the worst. The FFT model fares slightly better than the LoRA model. The \mbox{$\mathrm{RST^p_{w}}$-LoRA} model outperforms other neural summarization systems across all metrics, and its average scores on some  indicators approach the level of human performance. Moreover, compared to other neural summarization systems, the \mbox{$\mathrm{RST^p_{w}}$-LoRA} model is more likely to be recognized for producing the highest quality summaries and less likely to be considered as generating the poorest quality summaries.

\paragraph{GPT-4 Evaluation}

Inspired by \citet{liu-etal-2023-g}, we engage GPT-4 to assess our candidate models using the same guidelines as our human evaluators. To ensure experimental consistency, all experiments use the identical hyper-parameters settings detailed in Appendix \ref{sec:appendix1}. To avoid potential biases from previous interactions, we reset the conversation history prior to each query and abstain from making any further modifications. In our initial investigation, we aim to explore the extent to which GPT-4 evaluations\footnote{Utilizing the same iteration of the GPT-4 model as employed in prior summary generation tasks.} generally concur with human assessments in terms of both relative ranking and average scores within the same subset of 10 samples delineated in human evaluations. We then extend the evaluation to include all samples from the test sets\footnote{Prompt can be found in Appendix \ref{sec:appendix1_2}.}. 

The outcomes for these tests are shown in Table \ref{tab:gpt4_evaluation_BC}, as well as in Table \ref{tab:gpt4_evaluation_ML}, \ref{tab:gpt4_evaluation_eLife} in Appendix \ref{sec:appendix4}. We find that in GPT-4 evaluation, GPT-4 tends to assign the lowest scores to its own answers compared to those generated by other fine-tuned models. Summaries written by humans receive the highest scores and are generally regarded as the highest quality. In line with human evaluation findings, GPT-4 also recognizes LoRA as yielding inferior outcomes. In addition, the \mbox{$\mathrm{RST^p_{w}}$-LoRA} model scored higher than both LoRA and FFT. We further discuss the error analysis (case study) in Appendix \ref{sec:appendix3_1}.

\begin{table}[t]
\centering
\tabcolsep=3pt
\scalebox{0.85}{
\begin{threeparttable}
\begin{tabular}{lccccc}
\toprule
Candidate & R & I & C & F & Best $\mid$ Worst\\
\hline
Human & 4.89 & 4.76 & 4.67 & 4.72 & 96.8\% $\mid$ 0.0\% \\
\hdashline
GPT-4$_{ICL}$ & 4.02 & 3.81 & 4.47 & 3.12 & 0.0\% $\mid$ 35.3\% \\
Vicuna$_{LoRA}$ & 4.20 & 3.82 & 4.43 & 3.37 & 0.0\% $\mid$ 29.5\% \\
Vicuna$_{FFT}$ & 4.31 & 4.04 & 4.49 & 3.55 & 0.0\% $\mid$ 25.5\% \\
Vicuna\mbox{$_{RST^p_{w}-LoRA}$} & 4.46 & 4.44 & 4.60 & 4.12 & 3.2\% $\mid$ 9.7\% \\
\bottomrule
\end{tabular}
\end{threeparttable}
}
\caption{GPT-4 evaluation results on BC dataset}
\label{tab:gpt4_evaluation_BC}
\end{table}

\section{Conclusion}

We present RST-LoRA, a novel discourse-aware LoRA model tailored for long document summarization. Our approach primarily incorporates rhetorical knowledge into the LoRA training process by transforming RST structures into RST distributions. We develop four RST-LoRA variants, examining the impact of uncertainty in RST relational connections and discourse labels on overall performance. Empirical evidence from our studies demonstrates a consistent improvement in the performance of the standard LoRA model. By fine-tuning less than 0.5\% of the LLMs parameters, our best RST-LoRA variant not only surpasses the performance of LoRA and FFT but also exceeds previous state-of-the-art methods. Furthermore, our analysis underscores the efficacy of our approach in leveraging discourse knowledge, which strengthens LoRA's capabilities in producing more factually consistent and better-quality summaries.

\section{Ethics Considerations}
The datasets employed in our research are accessible to the public. Throughout the stages of data processing, experimental analysis, and model training/evaluation, our approach detects no violations of privacy. Regarding human evaluation, all participants engage voluntarily and are appropriately compensated. Additionally, we guarantee a safe and supportive setting during the evaluation period, following the ACM Code of Ethics in our experimentation and analysis.

\section{Limitations}

\paragraph{Data} All long document summarization datasets we use are open-source and peer-reviewed datasets. While these data sources are of high quality, inherent bias may exist within them. Exploring bias falls outside the scope of our study. In addition, the datasets we selected are from different fields (books, scientific publications, legal documents), and the heterogeneity between datasets is relatively high, but it is important to note that these data only represent a small fraction of real-world data and do not cover all possible long document summarization fields. Furthermore, although the RST parser we chose is multilingual, the exclusivity of English in our dataset can be seen as a limitation because it does not contain data for other languages, and we have not discussed RST-LoRA for non-English languages. 

\paragraph{Model} In our experiments, we employ two state-of-the-art long document pre-trained LLMs: Longformer and Vicuna. These models may carry biases from their pre-training phase. However, evaluating the extent of bias in these models has not been conducted, as it also lies outside the scope of this study. Additionally, the substantial cost of requesting GPT-4 (non-open source) API for generating summaries poses considerable financial barriers. We acknowledge this as a limitation and designate it as a potential space to explore in our future research. Furthermore, in our experiments, we compare the performance differences of different scales of LLMs (Longformer and Vicuna) and demonstrate that RST-LoRA can improve the summarization performance of LLMs at different scales. However, the number of potential LLMs is infinite, we have not tested on other models, and we leave the exploration of how the performance of RST-LoRA changes with different sizes of LLMs to future research.

\paragraph{Automated Evaluation} Although we apply a range of widely used automated evaluation metrics in our experiments to systematically assess candidate models from multiple perspectives on the test set. While these metrics provide a multi-faceted view of model performance, we are aware of their inherent limitations and the possibility that they may not fully encapsulate the models' comprehensive performance.

\paragraph{Parser} In line with prior research integrating discourse structures into summarization models, our work also needs an RST parser. In addition, manually annotated RST trees are extremely costly, we are unable to compare the summarization difference between RST parser output against human-annotated RST trees. Furthermore, incorporating RST into the LoRA training process does not significantly increase the amount of calculation or time complexity (just one more element-wise multiplication operation), but using a parser to generate the discourse structure still requires corresponding calculations. Our paper argues that discourse structure aids in summarization, which is orthogonal to the use of a specific discourse parser. We recognize that similar or greater improvements would be observed when employing a better parser.

\paragraph{Generalizability} Since the main research scope of this paper is long document summarization, we have not delved deeply into applying our proposed method to other NLP tasks, such as machine translation, question answering, or text simplification. Although our method could potentially be adapted to other NLP tasks involving LLMs without considerable modification, this aspect remains unexplored and is earmarked for subsequent research studies.

\paragraph{Human Assessment} The sample size for human evaluation in our study is constrained due to the nature of the extensive length of the original documents, which often extends over multiple pages. Scaling up the evaluation process through methods such as crowd-sourcing becomes challenging. Therefore, similar to many preceding studies \cite{atri-etal-2023-promoting, tang-etal-2023-improving, phang-etal-2023-investigating}, we evaluate only a set of 10 documents, which may not provide a fully representative view of the entire dataset. All human evaluators we recruit are Master's or Ph.D. students, not all are experts in the field of text summarization domain, nor are they necessarily proficient in reading across diverse domains. As such, their assessments, while valuable, should not be the sole basis for judgment.

\section*{Acknowledgements}
This project has received funding from the European Research Council (ERC) under the European Union’s Horizon 2020 Research and Innovation Programme (Grant Agreement No. 948878). We are grateful to the anonymous reviewers and area chairs for their exceptionally detailed and helpful feedback.
\begin{figure}[H] 
\centering
\includegraphics[width=0.8\columnwidth]{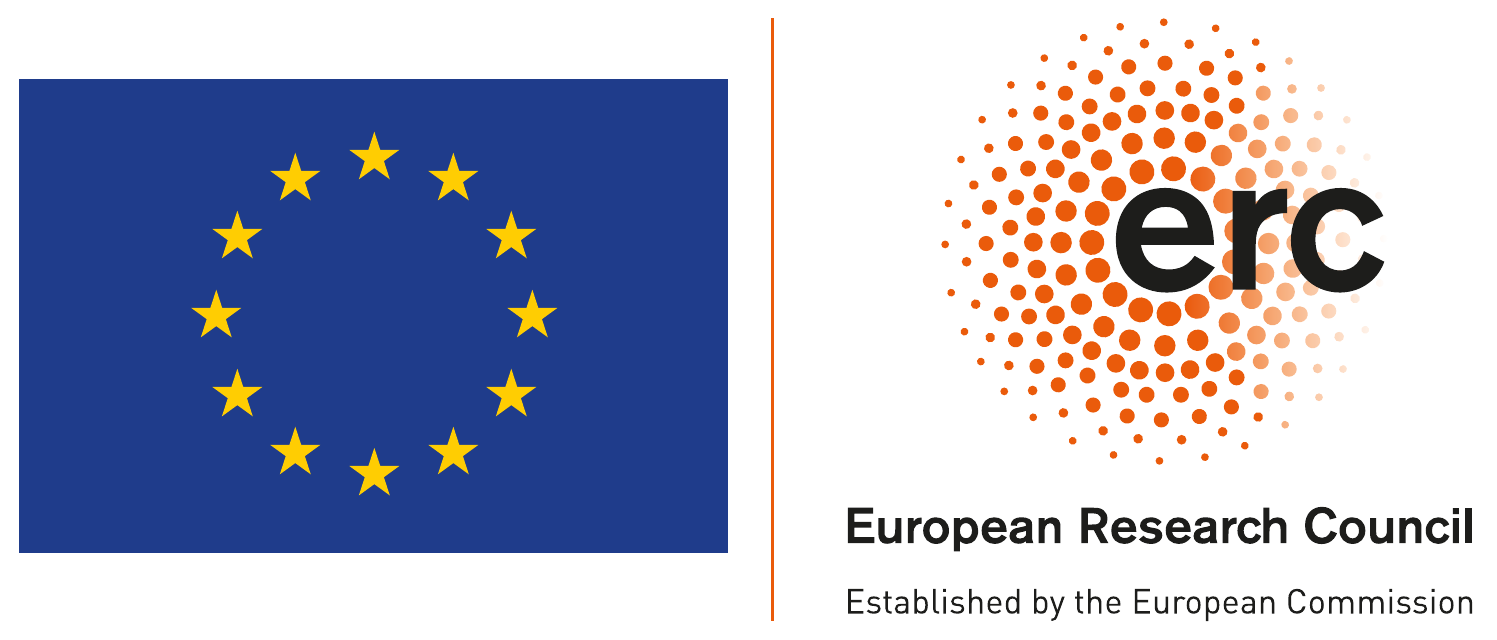}
\end{figure}

\bibliography{anthology,custom}

\appendix

\vspace{40pt}
\section{RST Relation Category}

\label{sec:appendix0}
\begin{table}[ht]
\centering
\scalebox{0.85}{
\begin{tabular}{c c}
\toprule
RST type & RST label \\
\midrule
\textit{Temporal} & Asynchronous, Synchronous\\
\textit{Contingency} & Cause, Condition\\
\textit{Comparison} & Contrast, Concession\\
\textit{Expansion} & Explanation, Elaboration, Conjunction\\
\bottomrule
\end{tabular}
}
\caption{RST relation category}
\label{tab:RST_relation_type}
\end{table}

\section{Datasets Statistics}
\label{sec:appendix0_1}

Table \ref{tab: datasets_discription} shows the statistics of the datasets. Within the table, \textit{Coverage} measures how much of the summary directly utilizes tokens from the source material. A higher coverage score suggests a greater proportion of the summary's tokens originate from the source document. \textit{Density} calculates the average length of source text segments each summary token is associated with. A higher density score could imply the inclusion of longer continuous text segments in the summary \cite{segarra-soriano-etal-2022-dacsa}. The \textit{compression ratio} represents the relationship between the source document's length and that of the summary. A lower compression ratio indicates a summary that is comparatively more concise.

\begin{table*}[ht]
\centering
\scalebox{0.8}{
\tabcolsep=4pt
\begin{tabular}{c c c c c c c c c}
\toprule
Dataset & Training & Validation & Test & Avg. Doc Tokens & Avg. Summary Tokens & Coverage & Density & Compression Ratio\\
\midrule
Multi-LexSum & 3177	& 454 & 908	& 75543.21 & 646.14 & 0.93 & 3.39 & 96.4\\
eLife & 4346 & 241 & 241 & 10132.12 & 382.58 & 0.82 & 1.77 & 27.65 \\
BookSum Chapter & 9600 & 1431 & 1484 & 5339.62 & 505.42 & 0.78 & 1.69 & 15.97 \\
\bottomrule
\end{tabular}
}
\caption{Datasets statistics}
\label{tab: datasets_discription}
\end{table*}

\section{Automatic Evaluation Metrics}
\label{sec:appendix0_2}
\begin{itemize}[leftmargin=8pt,itemsep=1pt,topsep=1pt,parsep=1pt]
    \item ROUGE \cite{lin-2004-rouge} evaluates the overlap of n-grams between the machine-generated summaries and human-crafted references. Our analysis includes the F1 scores for Rouge-1 (R1), Rouge-2 (R2), Rouge-L (RL), and Rouge-Lsum (RLsum).
    \item BERTScore \cite{zhang2019bertscore} uses BERT embeddings to analyze semantic similarity.
    \item METEOR \cite{banerjee-lavie-2005-meteor} computes the harmonic mean of uni-gram precision and recall, placing additional emphasis on recall for a balanced assessment.
    \item sacreBLEU \cite{post-2018-call} measures linguistic alignment and the fluidity between generated and reference summaries.
    \item NIST \cite{lin-hovy-2003-automatic} appraises the n-grams' informativeness by assigning weights based on the novelty of information they contain, as determined by their frequency in the corpus.
\end{itemize}

\section{Hyper-parameters Settings}
\label{sec:appendix1}
All experiments are optimized using the Adam \cite{kingma2014adam} optimizer (with $\beta_1$ = $0.9$, $\beta_2$ = $0.999$, $\epsilon$ = $10^{-9}$, and weight decay = $0.1$) and Adafactor \cite{shazeer2018adafactor}, with a warm-up ratio of 0.2. The initial learning rate is set to 5e-5, with a cosine learning rate schedule.

Additionally, within the LoRA strategy, we set a constant rank $r$ to 8, the scaling $\alpha$ to 32, and the dropout rate to 0.1. During training, we save checkpoints that achieve the highest Rouge-2 F1 score on the validation set as the final model. All experiments are run for 50 epochs with a batch size of 16, and early stopping is implemented to prevent over-fitting (all models converged before 50 epochs). For model inference, we employ a beam search of size 4 with a length penalty of 3.0 and set a no-repeat n-gram size of 3.

For GPT-4\footnote{\url{https://platform.openai.com/docs/models/}}, we employ GPT-4 Turbo version (\textit{gpt-4-1106-preview}), which is, at the time of experimentation (between
10 October 2023 and 15 December 2023), the best-performing publicly accessible version provided by OpenAI. For the hyper-parameters setting, we set temperature=1, top\_p=1, frequency penalty=0.2, and presence penalty=0.2. The remaining hyper-parameters are set to their default values as recommended by OpenAI.

\section{GPT-4 Prompts}
\subsection{Prompt for Zero-shot Summaries Generation}
\label{sec:appendix1_0}
\begin{mdframed}[backgroundcolor=cyan!4]
Document: \{Document\}\\
Summary:
\end{mdframed}

\subsection{Prompt for In-context Summaries Generation}
\label{sec:appendix1_1}
\begin{mdframed}[backgroundcolor=cyan!4]
Document: \{Document\} \\
Summary: \{Summary\} \\ \\
Document: \{Document\} \\ 
Summary: \{Summary\} \\ \\
Document: \{Document\}\\
Summary:
\end{mdframed}

\subsection{Prompt for Summaries Evaluation}
\label{sec:appendix1_2}
\begin{mdframed}[backgroundcolor=cyan!4]
Source Document: \{Document\} \\
Summary of Candidate1: \{Candidate1\} \\
Summary of Candidate2: \{Candidate2\} \\
Summary of Candidate3: \{Candidate3\} \\ 
Summary of Candidate4: \{Candidate4\} \\
Summary of Candidate5: \{Candidate5\} \\ \\
Note: The summaries are presented in order, with their respective candidate numbers from 1 to 5.\\ \\
Please review the following evaluation guidelines to assess the quality of the above five candidate summaries, and rank them from best to worst:\\ \\
Evaluation Guidelines: \{Guidelines\} \\ \\
Please use the following format for your output (scores ONLY):\\ \\
Relevance of Candidate1: \\
Informativeness of Candidate1:  \\
Conciseness of Candidate1: \\ 
Faithfulness of Candidate1: \\
Relevance of Candidate2: \\
Informativeness of Candidate2:  \\
Conciseness of Candidate2: \\ 
Faithfulness of Candidate2: \\
Relevance of Candidate3: \\
Informativeness of Candidate3:  \\
Conciseness of Candidate3: \\ 
Faithfulness of Candidate3: \\
Relevance of Candidate4: \\
Informativeness of Candidate4:  \\
Conciseness of Candidate4: \\ 
Faithfulness of Candidate4: \\
Relevance of Candidate5: \\
Informativeness of Candidate5:  \\
Conciseness of Candidate5: \\ 
Faithfulness of Candidate5: \\ 
Ranking (using candidate number): 
\end{mdframed}

\vspace{100pt}

\section{Impact of Different $r$}
\label{sec:appendix2}

\begin{figure}[h]
  \centering
  \includegraphics[width=0.47\textwidth]{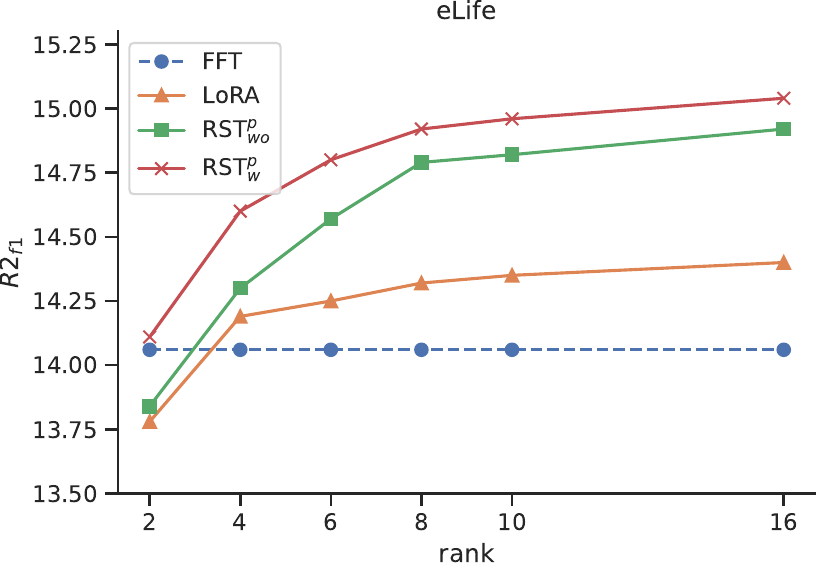}
  \caption{Impact of different $r$ on eLife dataset}
  \label{fig:diff_r_elife}
\end{figure}

\begin{figure}[h]
  \centering
  \includegraphics[width=0.47\textwidth]{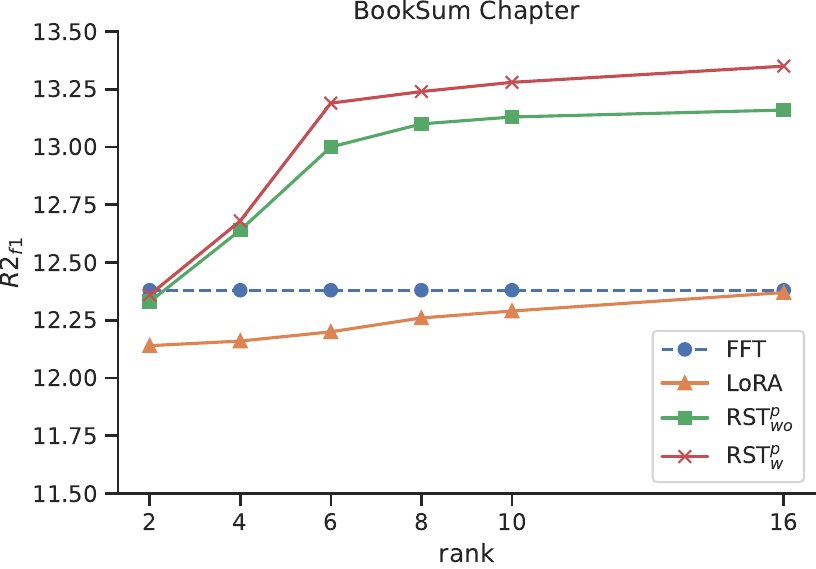}
  \caption{Impact of different $r$ on BC dataset}
  \label{fig:diff_r_BC}
\end{figure}

\FloatBarrier
\section{Human Evaluation Guideline}
\label{sec:appendix3}

Here, we offer a more detailed explanation of the metrics and evaluation criteria used for our human evaluation process.

\textbf{Prerequisites:} Eligibility for this evaluation requires simultaneous fulfillment of two conditions: (1) being a master's or Ph.D. student in Computer Science or Computational Linguistics, and (2) demonstrating greater than or equal to C2 English proficiency\footnote{\url{https://en.wikipedia.org/wiki/C2_Proficiency}}. If you do not meet both criteria, we respectfully ask you to refrain from participating in this task. Those who qualify are encouraged to proceed and follow the instructions below.

\vspace{60pt}

\begin{mdframed}[backgroundcolor=cyan!4]
We invite you to carefully review the following long document along with five candidate summaries. After a thorough examination of each summary, please rate them based on the following four criteria, using a Likert scale from 1 (worst) to 5 (best), where a higher score denotes better quality:

\begin{itemize}[leftmargin=8pt,itemsep=1pt,topsep=1pt,parsep=1pt]
    \item \textbf{Relevance:} This metric assesses the extent to which the summary content accurately reflects the source text. A relevant summary should encompass topics pertinent to the source document.
    \item \textbf{Informativeness:} This metric assesses the extent to which the summary provides a comprehensive understanding of the key points and essential details from the source text. An informative summary should encapsulate the core ideas, facilitating a clear and precise comprehension of the main arguments and findings of the source document.
    \item \textbf{Conciseness:} This metric assesses the extent to which the summary excludes less important information from the source text. A concise summary should effectively eliminate non-essential content from the source document during the generation process. 
    \item \textbf{Faithfulness:} This metric assesses the extent to which the candidate is incorrect in that it contradicts the information from the source document. A faithful summary adheres strictly to the information provided in the source document, avoiding the inclusion of unverified facts. 
\end{itemize}

Next, you are also expected to rank the candidates from best to worst based on overall quality.
\end{mdframed}

\vspace{500pt}

\section{GPT-4 Evaluation Results}
\label{sec:appendix4}

\begin{table}[!h]
\centering
\tabcolsep=2pt
\scalebox{0.85}{
\begin{threeparttable}
\begin{tabular}{lccccc}
\toprule
Candidate & R & I & C & F & Best $\mid$ Worst\\
\hline
Human & 4.67 & 4.70 & 4.52 & 4.83 & 94.2\% $\mid$ 0.0\% \\
\hdashline
GPT-4$_{ICL}$ & 4.43 & 3.88 & 3.62 & 3.19 & 0.0\% $\mid$ 43.3\% \\
Vicuna$_{LoRA}$ & 4.52 & 4.03 & 4.20 & 3.40 & 0.0\% $\mid$ 28.4\% \\
Vicuna$_{FFT}$ & 4.52 & 4.06 & 4.28 & 3.58 & 0.0\% $\mid$ 21.6\% \\
Vicuna\mbox{$_{RST^p_{w}-LoRA}$} & 4.57 & 4.33 & 4.31 & 4.22 & 5.8\% $\mid$ 6.7\% \\
\bottomrule
\end{tabular}
\end{threeparttable}
}
\caption{GPT-4 evaluation results on ML dataset}
\label{tab:gpt4_evaluation_ML}
\end{table}

\begin{table}[!h]
\centering
\tabcolsep=2pt
\scalebox{0.85}{
\begin{threeparttable}
\begin{tabular}{lccccc}
\toprule
Candidate & R & I & C & F & Best $\mid$ Worst\\
\hline
Human & 4.80 & 4.81 & 4.72 & 4.78 & 96.3\% $\mid$ 0.0\% \\
\hdashline
GPT-4$_{ICL}$ & 4.22 & 3.91 & 4.35 & 3.45 & 0.0\% $\mid$ 45.2\% \\
Vicuna$_{LoRA}$ & 4.47 & 4.12 & 4.41 & 3.58 & 0.0\% $\mid$ 30.1\% \\
Vicuna$_{FFT}$ & 4.59 & 4.23 & 4.47 & 3.82 & 0.2\% $\mid$ 16.3\% \\
Vicuna\mbox{$_{RST^p_{w}-LoRA}$} & 4.62 & 4.49 & 4.63 & 4.39 & 3.5\% $\mid$ 8.4\% \\
\bottomrule
\end{tabular}
\end{threeparttable}
}
\caption{GPT-4 evaluation results on eLife dataset}
\label{tab:gpt4_evaluation_eLife}
\end{table}

\section{Error Analysis}
\label{sec:appendix3_1}

The passage discusses a comparison between RST-LoRA and GPT-4 in generating summaries using two samples from the eLife test set, as shown in Tables \ref{tab:case_study_1} and \ref{tab:case_study_2}. It is observed that the summaries generated by GPT-4 tend to be overly general, lacking relevance and informativeness to the source document, and often include the model's own judgments or inferences. For instance, in the example provided, GPT-4 deviates from the main discussion topic and fails to align the generated summary with factual consistency from the source document. This comparison also aligns with the findings of \citet{atri-etal-2023-promoting} suggesting that general-purpose LLMs like GPT-4, without task-specific fine-tuning, still tend to focus more on language fluency but struggle to maintain fidelity to factual information and consistency with the source document.

\begin{table*}[t]
\centering
\scalebox{1}{\tabcolsep=3pt
\begin{threeparttable}
\begin{tabular}{|p{3cm}|p{12cm}|}
\toprule
\small Model & \small Output \\
\midrule
\small Target-1 & \small Imagine a gymnastics competition in which participants take turns to cartwheel and somersault across the floor. The routines on display comprise sequences of precisely timed movements learned through practice. This is also true for many of the actions we perform every day, such as reaching for a cup of coffee. A region of the brain called the cerebellum helps us learn sequences of movements. But how does it do this? To find out, Khilkevich et al. came up with a new version of an old experiment. Rabbits were first trained to blink their eye in response to a specific external cue. This type of learning, called associative learning, has been shown before in the cerebellum. But Khilkevich et al. wondered whether the cerebellum could also use internal feedback signals from the eye-blink as a cue to learn the next movement. If so, this might explain how the cerebellum can chain movements together in a sequence. As predicted, Khilkevich et al. found that rabbits could learn to blink their eye in response to an initial signal, and then blink again in response to the first blink. Control experiments confirmed that the second eye-blink was coupled to the first, and not to the original cue. Moreover, on many trials, the rabbits showed a third and even fourth eye-blink. This is because feedback signals from the first, second, or third blink were the same. Thus, the feedback signals from the first blink triggered the second blink, feedback from the second triggered the third, and so forth. Rabbits could also learn to use a blink of the left eye as a cue for a blink of the right eye. Similar patterns of neuronal activity accompanied each blink, suggesting that the same mechanism generated them all. The cerebellum can thus use feedback from one movement as a cue to learn the proper timing of the next movement in a sequence. A key question is whether this mechanism of sequence learning extends beyond movement. The cerebellum has extensive connections to the brain’s outer layer, the cortex, including many areas involved in cognition. Future experiments should test whether the cerebellum might help guide sequences of cortical activity during cognitive tasks.\\
\hdashline
\small GPT-4$_{ICL}$ & \small In a recent study, researchers explored the idea of using rabbits to understand human movement, focusing on how the cerebellum, a brain part, might be involved. They looked at how rabbits respond to various light and sound cues, seemingly to understand if these animals could learn complex dance-like moves. The study, which seemed to have a loose structure, suggested that the rabbits' performance in these tasks could somehow be linked to human learning of sequences, like playing a musical instrument or participating in sports. However, the connection between the rabbits' activities and human motor skills was not established, leaving the real impact of the study on human neuroscience quite ambiguous. Additionally, the research didn't delve deeply into the differences between rabbit and human brains, making its applicability to human subjects questionable.\\
\hdashline
\small Vicuna\mbox{$_{RST^p_{w}-LoRA}$} & \small The cerebellum's role in movement and learning was explored in a study by Khilkevich et al., where rabbits were trained to blink in response to cues. This research investigated the cerebellum's capacity for associative learning and its ability to use internal feedback from one movement as a cue for the next. Findings revealed that rabbits could learn sequential blinking, with each blink triggering the subsequent one. This suggests that the cerebellum not only coordinates simple movements but can also chain complex sequences of movements together. The study provides insights into the cerebellum's broader role in learning and cognition, particularly in understanding how it processes and coordinates the timing of sequential movements.\\
\bottomrule
\end{tabular}
\end{threeparttable}
}
\caption{Comparison of target summary with the summary generated by RST-LoRA and GPT-4}
\label{tab:case_study_1}
\end{table*}

\begin{table*}[t]
\centering
\scalebox{1}{
\tabcolsep=3pt
\begin{threeparttable}
\begin{tabular}{|p{3cm}|p{12cm}|}
\toprule
\small Model & \small Output \\
\midrule
\small Target-2 & \small To grow and multiply, a living cell must take a variety of factors into account, such as its own energy levels and the availability of nutrients. A protein called mTOR sits at the core of a signaling pathway that integrates these and other sources of information. Problems with the mTOR pathway contribute to several diseases including diabetes and cancer. The mTOR protein occurs in two distinct protein complexes, called mTORC1 and mTORC2. These complexes contain a mix of other proteins – known as accessory proteins. They also sense different cues and act upon distinct targets in the cell. Recent research reported the structure of mTORC1, which provided clues about how this complex works. Yet, much less was known about the mTORC2 complex. Stuttfeld, Aylett et al. have now used a technique called cryo-electron microscopy to reveal the three-dimensional architecture of the human version of mTORC2. Comparing the new mTORC2 structure to the existing one for mTORC1 showed that they have many features in common but important differences too. The overall shape of both complexes is similar and each complex contains two copies of mTOR arranged similarly. Also, the main accessory proteins in each complex interact with almost the same parts of mTOR, but the accessory proteins in mTORC2 are organized differently from those of mTORC1. The different accessory proteins also have distinct shapes. These differences could help to explain why the complexes respond to different cues and recognize different targets. These new findings provide an entry point for further studies on how mTORC2 works in cells. The next step is to get a higher-resolution image of the structure of this complex to see the finer details of all the components. This may in the future help scientists to develop drugs that inhibit mTORC2 to treat cancer and other diseases.\\
\hdashline
\small GPT-4$_{ICL}$ & \small In this paper, researchers delve into the intriguing world of cellular biology, focusing on a protein known as the mammalian target of rapamycin (mTOR). This protein, often likened to a cellular conductor, orchestrates a symphony of biological processes, guiding cells through complex growth and metabolic pathways. The study particularly highlights the dual roles of mTOR in different cellular complexes, drawing attention to its versatile nature. Interestingly, the paper touches upon the enigmatic behavior of mTOR in various diseases, hinting at its Jekyll and Hyde personality in cellular functioning. This study provides a glimpse into the intricate ballet of molecular interactions, painting a picture of mTOR as a pivotal figure in the dance of life.\\
\hdashline
\small Vicuna\mbox{$_{RST^p_{w}-LoRA}$} & \small The cells of animals, yeast, and other eukaryotes all contain a wide range of proteins that carry out different roles. A protein called mTOR sits at the core of this protein, where one letter of the protein sits at one end and the other at the other. The number of mRNAs is so small that it is thought that mTOR is only recently shown to be present in living cells. However, mTOR has only been studied in the laboratory in the form of single-celled structures called human cells. This has made it impractical for researchers to dissect the structure of the mTOR protein and its corresponding protein. Now, Wassagano et al. have used a technique called cryo-electron microscopy to reveal the three-dimensional architecture of mTOR proteins. The experiments reveal that this complex contains a small region called the ‘active’ site that is responsible for producing mTOR. This complex is also involved in the release of energy from blood sugar in cells and is also responsible for many of the processes that occur in the pancreas and immune responses in cells. In particular, the active state of the innermost mTOR complex is activated when the cell is starved for nutrients. Further experiments showed that the active site is involved in both the production and the activity of a protein called sIN1. Wassagoni et al al.'s findings show that both mTOR and SIN1 have a similar active site to each other in human cancers. These findings shed new light on the role of how mTOR interacts with other proteins in human cells, which may help to identify new targets for developing treatments against these kinds of cancer.\\
\bottomrule
\end{tabular}
\end{threeparttable}
}
\caption{Comparison of target summary with the summary generated by RST-LoRA and GPT-4}
\label{tab:case_study_2}
\end{table*}

\end{document}